# Visualizing Linguistic Shift

Salman Mahmood, Rami Al-Rfou and Klaus Mueller

**ABSTRACT**

Neural network based models are a very powerful tool for creating word embeddings, the objective of these models is to group similar words together. These embeddings have been used as features to improve results in various applications such as document classification, named entity recognition, etc. Neural language models are able to learn word representations which have been used to capture semantic shifts across time and geography. The objective of this paper is to first identify and then visualize how words change meaning in different text corpus. We will train a neural language model on texts from a diverse set of disciplines – philosophy, religion, fiction etc. Each text will alter the embeddings of the words to represent the meaning of the word inside that text. We will present a computational technique to detect words that exhibit significant linguistic shift in meaning and usage. We then use enhanced scatterplots and storyline visualization to visualize the linguistic shift

## 1 INTRODUCTION

The internet is a collection of materials produced by people from different time, geography, culture and disciplines. Therefore, the language on the internet demonstrates a rich variety of words and meanings. Words can gain new meanings which might be relevant to only a niche group of people. As the internet becomes more and more conversational, it is necessary to detect these linguistic changes.

Words can have more than one meanings; however, the context of the word can be used to determine the intended meaning of the word. For example, the word "fair" has multiple meanings. But in the sentence "It is not fair that I have to do all the work", the word fair means reasonable or just, whereas in the sentence "If you have fair skin you are more likely to get sunburnt", the meaning of the word fair is light or pale skin. In these sentences it is clear that the context in which the word is used can determine the meaning of the word. Therefore, the semantic distance between words can be determined by the context in which the word is used.

In order to find the semantic distance between words various models have been developed that use vectors to keep track of context in which the word was used, and then use different geometric techniques on the vectors to measure the similarity between words. The simplest method to achieve this is by creating a raw co-occurrence matrix of all the words that exist inside a window of a certain size. This method doesn't perform that well and better results can be achieved by using transformations on the data, such as reweighting the counts and using dimensionality reduction to smooth the data. These vector representations can then be used to calculate the linguistic shifts in the word. A lot of research has been directed towards linguistic shift, which we will discuss in section 2.

The objective of this paper is to use neural language models to find words that exhibit linguistic shifts in corpora belonging to different disciplines and then be able to visualize the linguistic shift. In order to find interesting words, we will be using neural language models to learn word embeddings. We propose a computational technique which allows us to discover words that have experienced significant change. In order to visualize the linguistic shift, we propose the use of storyline visualizations and scatterplots.

The structure of this paper is as follows in section 2 we will detail the related work, section 3 provides the problem statement, in section 4 we will present our method, section 5 discusses the different methods that can be used for visualization and finally section 6 provides the conclusion and future work.

## 2 RELATED WORK

Since our work lies at the intersection of different fields we will divide this section into four parts (1) language shift, (2) language shift visualization (3) word embeddings (4) storyline visualization.

### 2.1 Linguistic Shift

There is a large body of work that deals with how language changes, Bamman et al. [1] use Twitter data to study language shift in different genders, Bamman et al. [2, 11] also explore how geography and other contextual factors affect word meaning by using vector space representation of words. Einstein et al. [10] propose a model that identifies geographic and demographic factors that spread new words in online text. [8,12] present a model for measuring language change over time. In [18] Muhammad et al. present detailed study of distribution measures.

### 2.2 Linguistic Shift Visualization

Recently there has been an increase in the use of visual and dynamic applications in corpora and literature studies. Hilpert [6] uses dynamic motion charts to track the arrangements of the relative words, the sequence of graphs give the researcher an understanding of the complicated linguistic development. Motion charts are also used by others in this area as well [4, 5, 24]. Siirtola et al. [24] designed a tool, Text Variation Explorer (TVE), that allows interactive examination of the behavior of linguistic parameters affected by text window size and overlap, and performs interactive principal component analysis based on a user given set of words. Kulkarni et al. [12] use dimension reduction on their word vector and visualize them using a scatterplot.

### 2.3 Word Embeddings

The mapping of concepts to continuous space can be traced back to Hinton [Hinton] who proposed that concepts can be represented by distributed patterns of activity in networks on neuron like units. Bengio et al. [3] presented a neural language model for learning word embeddings, their model outperformed the traditional n-gram based techniques. Various methods have been proposed that scale and speed up large neural models [19, 23]. Mikolov et al. [15]

Salman Mahmood, Rami Al-Rfou and Klaus Mueller are with Stony Brook University, E-mail:{samahmood, ralrfou, mueller} @cs.stonybrook.edu

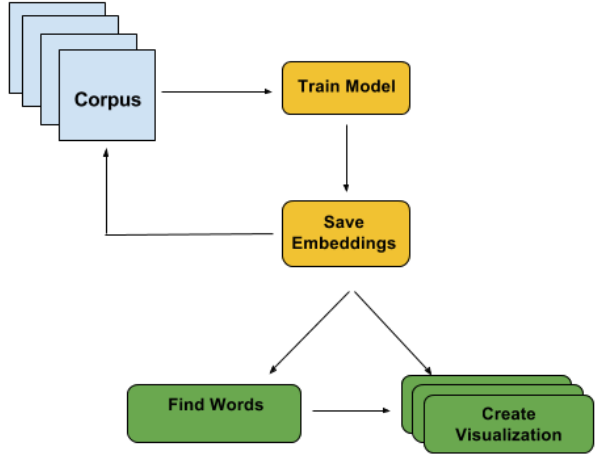

Figure 1: Shows an overview of the method.

proposed skip-gram models for learning word embeddings and demonstrated that these models have the capacity to learn linguistic patterns as linear relationships between vectors [16, 17]. These embeddings have been used in various Natural Language Processing tasks like such as named entity recognition (NER). Tsuboi et al. [27] incorporated word2vec embeddings with GloVe embeddings into a Part of Speech (POS) tagging system and show that it produces better results compared to the individual systems.

### 2.4 Storyline Visualization

Storyline visualizations have been inspired by the xkcd comic "Movie Narrative Charts" [20]. In a storyline visualization we visualize a collection of entities using line charts, where the lines converge if there is an interaction between the entities and diverge otherwise. Storyline visualizations have been used in visualization tools for wide variety of datasets. Vehlow et al. [28] use this technique to visualize dynamic community structure, by portraying communities as converged entities. Kim et al. [10] visualize genealogical data where individuals are represented using lines that converge and diverge to indicate marriage and divorce. Cui et al. [4] use text stream to visualize the evolution of topics, topics merge and split as they interact with each other.

A lot of research effort has been directed towards creating a generic visualization scheme, Ogievetsky et al. created an online editing tool, Plot Weaver [21], that allows users to create a storyline visualization by interactive editing. Automated schemes have also been produced by Tanahashi et al. [25] and Liu et al. [14], they have created schemes that produce results on par with professional artists. Further optimizations have been suggested by Kostitsyna et al. [11] and Tanahashi et al. [26] have developed a scheme to create storyline visualizations for streaming data.

### 3 PROBLEM STATEMENT

Our objective is to find words that exhibit significant shift in meaning across corpora from different disciplines. Given a corpora $C$, that contains texts from different fields $F$, for example, physics, law, religion etc, where corpus from each discipline is defined as $C_f$. We build a common vocabulary $V$ by intersecting the words in the corpus of each discipline $C_f$. For each word $w$ in $V$ we want to quantify how much that word changed meaning and be able to visualize how the word evolved across different corpus.

### 4 METHOD

An overview of the method is given in Figure 1, the method can be divided into three phases, we will first perform some data preprocessing, we then train the embeddings on each corpus using a neural language model and the final step is to identify words that exhibit the most shift in meaning.

### 4.1 Data Preprocessing

The text from the corpora is converted into sentences and all the punctuation is removed from the text. The initial training is done on a large dataset in order to make sure that there is a large number of words in model and that the embeddings of these words are well defined. We use the Wikipedia dataset to train the initial word2vec model.

### 4.2 Word Embeddings

Word2vec is a neural language model that can be used to train words in one of two ways, the first method predicts the word given the context, the model uses the preceding and following words therefore the input to the model is $wt - 2, wt - 1, wt + 1, wt + 2$ and the output of the neural network will be the word $wt$. This method is called continuous bag of words (CBOW). The second method predicts the context given the word, the model takes as input the word $wt$ and returns the context i.e. is $wt - 2, wt - 1, wt + 1, wt + 2$. This method is called skip-gram. We will be using the skip-gram model in our experiments because it is more suitable for large datasets, we use a window size of 5.

As shown in Figure 1 we train the model on the texts of corpus from different disciplines. We start by training the first model and saving the normalized word embeddings of the model, we repeat the process for all texts inside the corpora. We now have the embeddings of each training set. Here it should be mentioned that the order of the corpora might alter the results since we are building on the results of the previous corpus.

Given a corpora $C$, with text from different fields $F$ our objective is to learn field specific word embeddings $\phi_f : V, C_f \mapsto \mathbb{R}^d$. Where $V$ is vocaubulary of the corpora, and $C_f$ represents the corpus of the field $f$. Our objective is to find a word representations that can be used to predict the context words. Given a word $w_i$, the context words are defined as the words that appear before or after $w_i$ within a window of size $n$. We model the probability of a context word $w_j$ given $w_i$ as:

$$\Pr(w_j \mid w_i) = \frac{\exp(W_j^T w_i)}{\sum_{w_k \in V} \exp(W_j^T w_i)} \qquad (1)$$

here $w_i$ is the vector representation of $w_i$. To train the model we iterate over each word occurrence of $C_f$ to minimize the negative log-likelihood of the context word. The objective function is defined as follows:

$$J = \sum_{w_i \in C_f} \sum_{\substack{-n < j < n \\ j \,!= 0}} -\log \Pr(w_j \mid w_i) \qquad (2)$$

Here n is the size of the context window. This formulation is proportional to the size of $|V|$, which is generally very large. We optimize the process by mapping the problem from a classification of 1-out-of-V to a hierarchical classification problem [19]. This reduces the complexity from $O(|V|)$ to $O(\log|V|)$.

### 4.3 Distance Metric

The word embeddings that were learned are then used to find words that exhibit maximum linguistic change, in order to find these words, we use an ensemble technique in which we use the sum of two distance metrics, step-wise Euclidean distance and the nearest neighbor distance.

The step-wise Euclidean distance is calculated by finding the distance moved by the word in the embeddings of each model and this distance is normalized by using the log of the word count in each model. To be more concrete for a word $w$ the step-wise Euclidean distance is defined as follows:

$$EUC(w) = \sum_{f \in F} \frac{euc(Emb_f^w, Emb_{f-1}^w)}{\log(wc_w^f)} \quad (3)$$

where $EUC(w)$ is the step-wise Euclidean distance, $Emb_f^w$ is the embedding of the word $w$ in field $f$. $euc(a,b)$ is the Euclidean distance between vectors a and b, and $wc_w^f$ is the word count of $w$ in $f$.

Words that have similar context tend to have similar meaning, therefore a change in the nearest neighbors of a word is a good indicator of a shift in the meaning of a word. The nearest neighbor distance is the difference between the nearest neighbors of two words, summed over all embeddings. More precisely the nearest neighbor distance is defined as follows:

$$NN(w) = \frac{\sum_{e \in E} K - |nn_K^{w,e} \cap nn_K^{w,e-1}|}{\log(wc_w^f)} \quad (4)$$

Where $K$ is the number of nearest neighbours, $nn_k^{w,e-1}$ is defined as the $K$ nearest neighbors of the word $w$ in the embeddings e, and $wc_w^f$ is the word count of $w$ in $f$. In our experiments we used K =20.

|  | Wikipedia | Fiction | Religious | Politics |
|---|---|---|---|---|
| people | person | folk | Israel | disbelievers |
| life | upbringing | grief | toil | satisfaction |
| nakedness | malevolence | loveliness | unfaithful | unfaithful |
| time | period | moment | day | day |

Table 1: Shows a list of words and their nearest neighbours from the embeddings trained on different texts.

### 5 VISUALIZATION

Visualizing linguistic shift can be challenging, simply observing the change in embeddings of a word does not suffice. In order to provide a frame of reference we will use neighboring words at each segment. The neighboring words will have a similar context and hence they can provide an understanding of what the word means at that point. In order to visualize the linguistic shifts in the data we will be using enhanced scatterplots and storyline visualization.

In this section we will explain the methodology used to create these visualizations, in order to create these visualization, we trained our model on corpus from four different fields, Wikipedia, Fiction (Lord of the Rings trilogy), Religious (the King James version) and Politics (A Preface to Politics by Walter Lippmann). We used these texts because they represent a very diverse group and it will be interesting to see how words change meaning inside these texts. These corpora are used to train the model and find the words that exhibit most linguistic shift, as described in section 4, Table 1 shows some of the results of our method, it shows a list of words and their nearest neighbour from the embeddings trained on different texts. We will demonstrate the visualizations using the word "heart" and show how different visualization methods can be used to observe how the word changes meaning in texts from different fields.

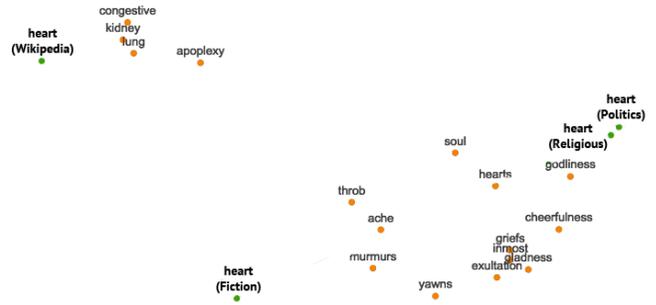

Figure 2: Shows the scatterplot visualization of the linguistic change of the word heart.

### 5.1 Enhanced Scatterplot

The scatterplot shows how the word heart changes position in different embeddings. We use the nearest neighbors of the words to identify the meaning of the word. Since the embedding of the neighboring words also change we use the mean of embeddings across different sections. Then we apply a dimension reduction technique to reduce the number of dimensions to 2 and visualize the graph as a scatterplot (see Figure 2). The figure shows the word "heart" as it is affected by the text of different corpus. We can see that the word heart is closer to biological organs in Wikipedia corpus. In fictional works its meaning is closer to words representing pain like ache and throb. In the religious and political texts, the meaning is closer to feelings like cheerfulness and grief. The visualization is able to show how the word heart changes meanings across different corpus. The limitation of this visualization is that the words are close together which makes it a little difficult to determine how the words are grouped in the embeddings. Also since the dimensionality of the embeddings is generally very large, a lot of information is lost in the dimension reduction step.

### 5.2 Storyline Visualization

Storyline visualization is a technique that has been inspired by the illustration of Munroe [20], the visualization shows the interaction between various characters in The Lord of the Rings movie. Lines that are clustered together depict the spatial proximity of the characters. Storyline visualization is an effective way to convey how relationships between different entities change over time. Storyline visualization fits nicely into our scheme since it allows us to show how the relationship between different words evolves over different segments. We believe that Storyline Visualization is an effective visualization technique for this scenario because even though it doesn't convey the exact location of the words but it very effectively conveys how the words are grouped together. It will allow us to determine if the word has moved to a different neighborhood and also show the entire evolution of the word.

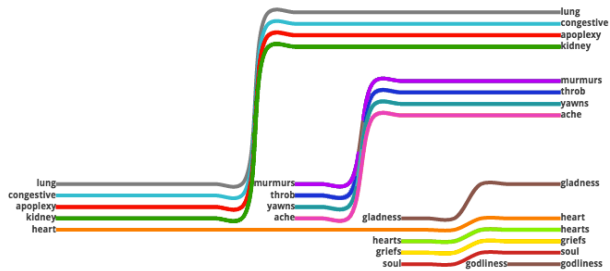

Figure 3: Shows the storyline visualization of the linguistic change of the word heart.

The algorithm to draw storyline visualization attempts to arrange the lines in a way that minimizes the number of crossings of the lines. The method we use to generate storylines is inspired by [26], however their method is made primarily for streaming data and one of their main concerns is the conservation of the user's mental mapping during the constant layout updates. Therefore, their layout modification is based on only the previous time steps. Since we are not using streaming data we will use all the data available to create our layout.

In order to create the visualization for a given word we need to cluster words together. To find these clusters we find the K nearest neighbors of the target word at each embedding. Out of the K nearest neighbors we pick the top $M$ words (where $M < K$) and cluster them together with the given word. To this cluster we then add any word in the top $M$ words of the previous cluster which are also in the top $K$ words of this segment. In the example shown we have used M =4 and K = 32.

Figure 3 shows the linguistic shift of the word "heart" and the groupings of the words are much more apparent in this visualization as compared to the scatterplot, very clearly conveying the meaning of the word in each embedding. However, the information related to the distances between the words is lost, making it difficult to tell which word is the closest match.

## 6 CONCLUSION AND FUTURE WORK

In this paper we have presented a method that can be used to find words that exhibit linguistic shift and we evaluated visualization techniques that can be used to visualize the linguistic shift of words. We show that the scatterplot visualization can be augmented with the storyline visualization to get more insight into the data. In the future we intend to create a comprehensive tool which can be used to analyse linguistic change in data. We will use some of the visualization techniques mentioned in this paper as well as provide other data, such Part of Speech tags, word frequency etc. that is commonly used in analysing linguistic change.